\documentclass[11pt]{article}

\usepackage[final]{acl}

\usepackage{times}
\usepackage{latexsym}

\usepackage[T1]{fontenc}

\usepackage[utf8]{inputenc}

\usepackage{microtype}

\usepackage{inconsolata}

\usepackage{graphicx}
\usepackage{booktabs}
\usepackage{amsmath} 
\usepackage{amssymb}
\usepackage{amsthm}
\theoremstyle{definition}
\newtheorem{definition}{Definition}
\usepackage{tcolorbox}
\usepackage{cuted}  
\usepackage{listings}
\usepackage{subcaption} 
\usepackage{algorithm}
\usepackage{algorithmic}

\definecolor{lightblue}{rgb}{0.933,0.968,0.988}

\title{Unearthing Gems from Stones: Policy Optimization with \\Negative Sample Augmentation for LLM Reasoning}

\author{
  \textbf{Zhaohui Yang\textsuperscript{1,3}\thanks{Equal contribution.}},
  \textbf{Yuxiao Ye\textsuperscript{2,3}\footnotemark[1]},
  \textbf{Shilei Jiang\textsuperscript{3}},
  \textbf{Chen Hu\textsuperscript{3}},
  \textbf{Shihong Deng\textsuperscript{3}\thanks{Corresponding author.}},
\\
  \textbf{Linjing Li\textsuperscript{1}},
  \textbf{Daxin Jiang\textsuperscript{3}},
\\
  \textsuperscript{1}Institute of Automation, Chinese Academy of Sciences\\
  \textsuperscript{2}Beijing Institute of Technology \ \textsuperscript{3}StepFun, China
\\
\texttt{yangzhaohui2023@ia.ac.cn} \quad \texttt{yuxiaoye99cs@gmail.com} \\
\texttt{dshnightmare@gmail.com}
}
\begin{document}
\maketitle
\begin{abstract}
Recent advances in reasoning language models have witnessed a paradigm shift from short to long CoT pattern. Given the substantial computational cost of rollouts in long CoT models, maximizing the utility of fixed training datasets becomes crucial.
Our analysis reveals that negative responses contain valuable components such as self-reflection and error-correction steps, yet primary existing methods either completely discard negative samples (RFT) or apply equal penalization across all tokens (RL), failing to leverage these potential learning signals. In light of this, we propose Behavior Constrained Policy Gradient with Negative Sample Augmentation (BCPG-NSA), a fine-grained offline RL framework that encompasses three stages: 1) sample segmentation, 2) consensus-based step correctness assessment combining LLM and PRM judgers, and 3) policy optimization with NSA designed to effectively mine positive steps within negative samples. Experimental results show that BCPG-NSA outperforms baselines on several challenging math/coding reasoning benchmarks using the same training dataset, achieving improved sample efficiency and demonstrating robustness and scalability when extended to multiple iterations.

\end{abstract}

\section{Introduction}

Reasoning capabilities are a critical aspect of evaluating the intelligence of large language models (LLMs). Recent advances have witnessed a paradigm shift from short to long chain-of-thought (CoT) reasoning, particularly after the release of OpenAI's o1 series \cite{openai2024learning} and Deepseek R1 \cite{deepseek-r1}.
However, generating lengthy CoT responses incurs substantial computational costs. In the online learning paradigm, the periodic rollout process has become the bottleneck of the post-training system \cite{deepcoder}, significantly reducing training efficiency. Consequently, maximizing the utility of fixed training datasets under offline learning paradigm becomes increasingly crucial.

Motivated by this, a natural question arises: \textit{How can we better utilize negative samples, particularly those from long CoT model?}
Currently, self-improvement algorithms for enhancing LLMs' reasoning capabilities fall into two categories: rejection sampling fine-tuning (RFT \cite{zhang2406rest, tian2025deepdistill}) and reinforcement learning (RL \cite{rafailov2023direct, topr, kimi_k1p5}), yet neither fully exploits the potential of negative samples.
For RFT methods, negative samples are entirely discarded.
While RL methods like DPO \cite{rafailov2023direct}, GRPO \cite{shao2024deepseekmath}, and GPG \cite{chu2025gpg} attempt to leverage negative samples, they simply apply equal penalization to all tokens without fine-grained discrimination.

We argue that while treating all steps in negative samples as flawed is reasonable for short CoT responses due to their brevity and directness, the situation differs significantly in long reasoning trajectories. Even when the final answer is incorrect, many intermediate steps can be valuable \cite{li2025llms, ahmad2025opencodereasoning}. As illustrated in Figure \ref{fig:method}(a), we observe that model responses often exhibit intrinsic behaviors such as verification, error-correction, and self-reflection \cite{min2024imitate, ahmad2025opencodereasoning}, with these patterns occurring more frequently than in positive samples, partly due to the typically longer response length of negative samples \cite{fatemi2025concise}. 
Therefore, in the long CoT pattern, simply rejecting all steps of negative samples is not sound, potentially undermining beneficial reasoning steps. 
PPO \cite{ppo} attempts token-level credit assignment through value model. Nevertheless, its effectiveness is limited by the exponentially growing action space under the long CoT paradigm, requiring substantial data for accurate token-level value estimation.

To better utilize negative samples, we propose a fine-grained offline RL framework named Behavior Constrained Policy Gradient with Negative Sample Augmentation (BCPG-NSA). 
Our framework first employs a semantic segmentation model to divide negative samples into multiple steps. These steps are then evaluated for correctness using an LLM judger and/or a process reward model (PRM) \cite{zheng2024processbench}.
Finally, we introduce a novel token-level policy optimization objective that enables fine-grained leveraging of negative samples, where the penalization for valuable steps is reduced or even reversed to encourage their generation, with this adjustment strength controlled by the mining coefficient.

Experimental results demonstrate that BCPG-NSA achieves the best performance and improved sample efficiency compared to baselines on challenging math reasoning benchmarks AIME24 and AIME25, and o.o.d coding benchmark LiveCodeBench.
In addition, our ablation studies show that the consensus-based LLM-PRM annotation approach yields the best results compared to LLM-only and PRM-only approaches, indicating the importance of precise step correctness assessment in negative samples.
We also demonstrate that BCPG-NSA is robust across different mining coefficient values, maintains stability over extended training epochs, and scales effectively to multiple iterations.

Our contributions are summarized as follows:
\begin{itemize}
    \item To the best of our knowledge, we are among the first to empirically validate the value of negative samples through experimental analysis and case studies, and propose mining correct reasoning steps from these samples to enhance long CoT reasoning.
    \item We introduce BCPG-NSA, an effective offline RL training framework that integrates reasoning step segmentation, consensus-based LLM-PRM annotation, and a policy optimization objective with negative sample augmentation. Across several challenging math and coding benchmarks, BCPG-NSA achieves improved performance and sample efficiency compared with baselines.
    \item We conduct extensive analyses to demonstrate the robustness and scalability of BCPG-NSA, providing insights into its effectiveness under various conditions.
\end{itemize}

\section{Related Work}
\textbf{Long Chain-of-Thought Reasoning Language Models} \ 
LLMs have demonstrated remarkable reasoning capabilities in complex tasks. CoT is one of the significant methods to encourage and enhance the reasoning ability of LLMs, which guides LLMs to break down problems and solve them step by step. 
OpenAI o1 \cite{openai2024learning} is the first to introduce inference time scaling law, which employs large-scale RL to enable autonomous optimization of CoT during training and overcome challenging tasks by generating more reasoning tokens. 
Several efforts \cite{qwq, deepseek-r1, kimi_k1p5, zhang2025srpo, hu2025open} have successfully replicated the inference time scaling law, demonstrating the powerful capabilities of this new inference paradigm. 

\textbf{Process Reward Models in Mathematical Reasoning} \ 
Mathematical reasoning in LLMs has made significant strides with the introduction of reward models. Reward models are primarily divided into two categories: Outcome Reward Model (ORM) and Process Reward Model (PRM). ORMs only score the final answer of the LLMs' responses, while PRMs assign scores to each reasoning step, providing granular feedback. Consequently, PRMs can not only guide search \cite{park2024ensembling, zhang2406rest} but also offer dense rewards in RL training \cite{gao2024designing}.

\textbf{Benefits from Negative Data} \ 
Many RL methods for LLMs, such as GRPO \cite{shao2024deepseekmath} and GPG \cite{chu2025gpg}, consider every step in a negative sample incorrect, and use the same strength to push down the likelihood of all tokens in incorrect responses. 
Prior research on learning from negative samples primarily focuses on training data construction. Recent works on DPO \cite{rafailov2023direct, setlur2024rl} propose constructing training pairs with shared prefixes between positive and negative samples, aiming to improve the model's decision-making at critical intermediate steps.
Another research targeting SFT data construction \cite{wang2024learning} additionally adds a prefix to indicate whether the current generation is a successful trajectory, helping the model better distinguish between correct and incorrect responses.
The most closely related work \cite{li2024turning} leverages valuable signals from negative samples in short CoT reasoning, through a specialized dual-LoRA framework for model distillation.
Under the long CoT paradigm, incorrect and correct steps more frequently alternate throughout the reasoning process. Therefore, fine-grained mining of the value of negative samples is a promising direction.

\section{Preliminary Analysis: The Value of Negative Samples}

In this section, we investigate two fundamental questions: \textit{1) Can supervised fine-tuning on negative samples bring performance benefits?} and if so, \textit{2) To what extent can these benefits be realized?}

Specifically, we construct two distinct training datasets from the open-source R1 dataset \cite{yang2025deepcritic}: \texttt{SFT-pos} and \texttt{SFT-neg}, to compare their respective performance gains when training the base model Qwen2.5-14B-Base \cite{qwen2.5}. Both datasets share identical prompts and contain an equal number of samples (19,000 each). The key difference lies in their composition: \texttt{SFT-pos} consists exclusively of responses with correct final answers, while \texttt{SFT-neg} contains only responses with incorrect answers.

\begin{table}[htbp]
    \centering
\begin{tabular}{@{}ccc@{}}
\toprule
                     & AIME24 & AIME25 \\ \midrule
Qwen2.5-14B-Instruct & 10.00  & 13.33  \\
- trained on \texttt{SFT-pos} & 52.75  & 39.42  \\
- trained on \texttt{SFT-neg} & 41.67  & 34.00  \\ \bottomrule
\end{tabular}
\caption{Performance improvement under different training set curations.}
\label{tab:sft_neg}
\end{table}

\begin{figure*}
    \centering
    \includegraphics[width=1\linewidth]{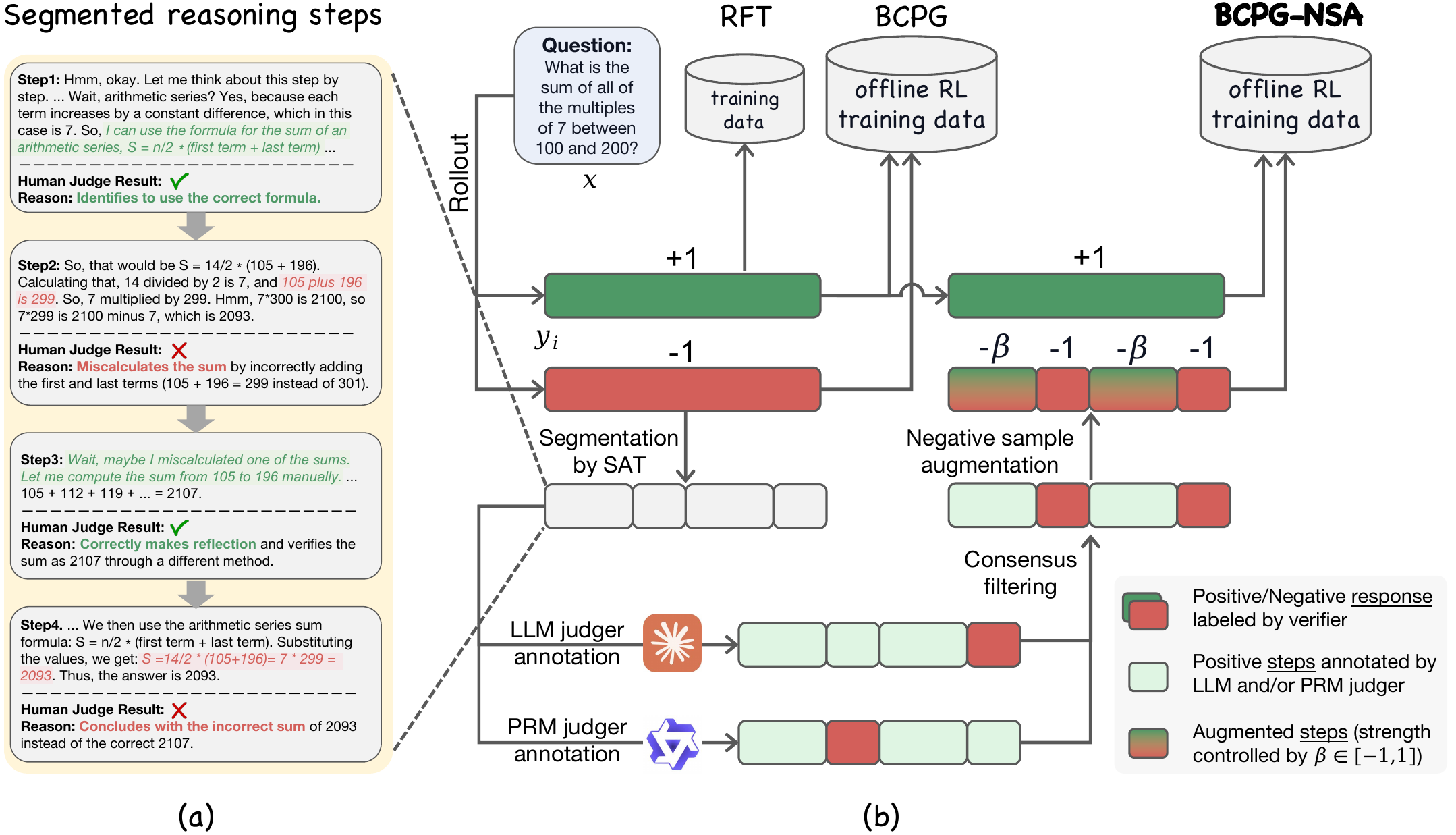}
    \caption{(a): A case study: presence of correct steps (via human judgments) within an incorrect response. (b): The overall framework of BCPG-NSA.}
    \label{fig:method}
\end{figure*}

We draw two conclusions from the results presented in Table \ref{tab:sft_neg}.
Firstly, fine-tuning on negative samples yields substantial performance gains, with the \texttt{SFT-neg} model outperforming the base Qwen2.5-14B-Instruct by 31.67\% on AIME24. This suggests the presence of valuable components within incorrect responses. Through human investigation, we find that long CoT models often exhibit self-reflection and propose alternative problem-solving approaches during reasoning. We illustrate a representative example in Figure \ref{fig:method} (with complete details provided in Section \ref{sec:case}).

Secondly, while \texttt{SFT-neg} significantly improves upon the base model, it still underperforms compared to \texttt{SFT-pos} by 5-10\% across benchmarks, due to the presence of flawed reasoning steps in negative samples. This observation motivates us to design a mechanism that efficiently distinguishes and leverages valuable steps within negative samples to maximize performance gains.

\section{Method}
Our method BCPG-NSA encompasses three stages: thinking process segmentation, consensus-based step correctness annotation, and policy optimization with negative sample augmentation. The overall framework of BCPG-NSA is shown in Figure \ref{fig:method}(b).
\subsection{Segmenting Thinking Process into Steps}

\label{sec:segmentation}
To enable a more fine-grained analysis of the negative sample thinking process, we first need to segment the thinking process into multiple steps. Previous approaches to segment the CoT process mainly fall into two categories: rule-based methods, such as splitting using double line breaks \cite{zhang2406rest}, and automatic segmentation using LLMs \cite{zhang2025lessons, zheng2024processbench}. Rule-based methods can lead to semantic discontinuity, such as incomplete reasoning steps or mixing multiple independent logical segments within a single step. On the other hand, automatic segmentation with LLMs may result in content loss after segmentation. Therefore, we choose to use the SAT model \cite{frohmann2024segment}. The SAT model not only automatically segments text based on semantics, but also ensures consistency of the text content before and after segmentation. We use a binary search method to find an appropriate segmentation threshold and make the number of steps after segmentation within a reasonable range.

Specifically, for each prompt $x$, we use a fixed reference policy $\pi_{\rm ref}$ to generate $G$ responses $\{y_i\}_{i=1}^G$. The rollout process of $\pi_{\rm ref}$ that auto-regressively generates the $i$-th response $y_i$ can be formulated as:
\begin{equation}\label{eqn:roll}
\pi_{\rm ref}(y_i|x) = \prod_{j=1}^{|y_i|} \pi_{\rm ref}(y_{i,j} | x,y_{i, <j}),
\end{equation}
where $y_{i,j}$ denotes the j-th token in response $y_i$.
Given the ground truth $y^*$, the correctness of $y_i$ is labeled by a verifier (e.g., the well-established \texttt{math\_verify} \cite{mathverify2025} for math reasoning tasks), yielding a binary reward $r_i = r(x, y_i, y^*) \in \{0, 1\}$. 
For negative samples (where $r_i = 0$), each response $y_i$ is segmented by the SAT model into $K$ consecutive steps:
\begin{equation}\label{eqn:seg}
\begin{array}{c}
y_i  \overset{\text{SAT}}{=} {\rm STEP}_{i,1} || {\rm STEP}_{i,2} || \cdots || {\rm STEP}_{i,K}, \\[7pt]
{\rm STEP}_{i,k}  \triangleq y_{i, {\rm start}_k} || \cdots || y_{i, {\rm end}_k},
\end{array}
\end{equation}
where $||$ is the concatenation operator, and ${\rm start}_k$ and ${\rm end}_k$ represent the indices of the starting and ending tokens of the $k$-th step, respectively.

\subsection{Consensus-based Annotation by LLM and PRM Judgers}
\label{sec:annotation}

After performing segmentation on the negative samples, we use an LLM judger and a discriminative PRM to jointly annotate each step in the reasoning process as correct/incorrect. 

In designing the LLM judger, we first establish a taxonomy of \textit{context-agnostic} reasoning errors, encompassing calculation mistakes, derivation errors, logical flaws, problem misinterpretations, and similar issues. 
However, our preliminary experiments reveal that these context-agnostic rules alone are insufficient for comprehensive evaluation. This limitation stems from the distinctive characteristic of long CoT LLMs: their capability to engage in self-reflection and error-correction during the reasoning process, often resulting in alternating sequences of correct and incorrect steps. Consequently, the assessment of a step's correctness cannot be performed independently, but rather depends on its relationship with the preceding context.

Therefore, we augment the annotation criteria with two \textit{context-aware} rules:
\begin{definition}[Error Propagation]
A reasoning step is classified as \textit{incorrect} if it satisfies two conditions: 1) it follows an incorrect step, or 2) it continues the reasoning based on the previous step without introducing new problem-solving approaches.
\end{definition}
\begin{definition}[Error Termination]
A reasoning step is classified as \textit{correct} if it follows an incorrect step and either: 1) successfully rectifies the previous error, or 2) introduces an alternative problem-solving approach.
\end{definition}
Formally, we inject these annotation criteria into the prompt template (see Figure \ref{fig:prompt_template_judger}) and instruct the LLM judger $\phi^{\rm LLM}(\cdot)$ to assess the correctness of each step:
\begin{equation}\label{eqn:llm-judger}
I_{i,k}^{\rm LLM} = \phi^{\rm LLM}(x, {\rm STEP}_{i,:k}) \in \{0, 1\},
\end{equation}
where $I_{i,k}^{\rm LLM} = 1$ denotes ${\rm STEP}_{i,k}$ is annotated as correct.

Similarly, we use a discriminative PRM $\phi^{\rm PRM}(\cdot)$ to annotate each reasoning step. The PRM first predicts the score $\sigma_{i,k}$ for ${\rm STEP}_{i,k}$, and then applies a threshold $\lambda$ to determine the annotation outcome: 
\begin{equation}\label{eqn:prm-judger}
\begin{array}{c}
\sigma_{i,k} = \phi^{\rm PRM}(x, {\rm STEP}_{i,:k}) \in [0, 1], \\[7pt]
I_{i,k}^{\rm PRM} = \mathbf{1}_{\sigma_{i,k}>\lambda} \in \{0,1\},
\end{array}
\end{equation}
where $\mathbf{1}(\cdot)$ is the indicator function. In our experiments, $\lambda$ is determined through grid search, with the selection criterion being to maximize the consistency between LLM and PRM annotating results.

Finally, we introduce a consensus filtering approach, in which a reasoning step is considered correct only when both the LLM judger and PRM deem it correct:
\begin{equation}\label{eqn:consensus}
I_{i,k} = I_{i,k}^{\rm LLM} \land I_{i,k}^{\rm PRM}.
\end{equation}

\subsection{Policy Optimization with Negative Sample Augmentation}
Drawing from classical offline RL work \cite{park2024value}, we propose a novel objective function that combines policy improvement with behavior constraint (i.e., KL regularization), while introducing a mechanism to leverage valuable components within negative samples. Our objective function BCPG-NSA is formulated as follows:
\begin{align}
    \label{eqn:obj}
    &J_{\rm BCPG-NSA}(\pi_\theta) = \notag \\
    &\frac{1}{G}\sum_{i=1}^{G}\frac{1}{|y_i|}\Bigg[\sum_{j=1}^{|y_i|}\mathrm{log}\pi_{\theta}(y_{i,j}|x, y_{i, <j})\beta_{i,j}(r_i-\overline{r}) \notag \\ 
    &\qquad\qquad\;\;\;\;\; - \frac{\tau}{2}\big(\mathrm{log}\frac{\pi_{\theta}(y_i|x)}{\pi_{\mathrm{ref}}(y_i|x)}\big)^2\Bigg],
\end{align}
where $\overline{r} = {\rm mean}(\{r_i\}_{i=1}^G)$ is the average reward for the group, $\tau$ is the behavior constraint factor. Notably, the value of $\beta_{i,j}$ is given by
\begin{equation}
    \beta_{i,j} =
    \begin{cases}
        \beta & \text{if } y_{i,j} \in {\rm STEP}_{i,k} \; \text{and } I_{i,k} = 1 \\
        & \phantom{\text{if }} \text{and } r_i=0 \\
        1, & \text{otherwise},
    \end{cases}
\end{equation}
where the ``if'' condition indicates that token $y_{i,j}$ appears in a correct step within an incorrect response.
The mining coefficient $\beta \in [-1, 1]$ is a hyperparameter that controls the degree of augmentation for correct steps within negative samples.
The smaller values of $\beta$ indicate stronger augmentation of correct steps in negative samples, and intuitively, the penalization for valuable tokens is reduced (when $\beta \in [0,1)$) or even reversed to encourage their generation (when $\beta < 0$).
Specifically, when $\beta=1$, BCPG-NSA reduces to vanilla BCPG, which penalizes all tokens in negative samples equally. In this case, the objective function resembles the loss formulation of Kimi k1.5 \cite{kimi_k1p5}. A detailed procedure of BCPG-NSA is illustrated in Algorithm \ref{alg:bcpg}.

\begin{algorithm}[H]
\caption{BCPG-NSA (single iteration)}\label{alg:bcpg}
\begin{algorithmic}
\STATE \textbf{Input:} Reference model $\pi_{\rm ref}$
\STATE /* Offline training data construction */
\FOR{$x$ in prompt set}
    \STATE Rollout by Equation~(\ref{eqn:roll})
    \IF{$r_i=0$}
        \STATE Response segmentation by Equation~(\ref{eqn:seg})
        \STATE Step annotation by Equation~(\ref{eqn:llm-judger})~(\ref{eqn:prm-judger})~(\ref{eqn:consensus})
    \ENDIF
\ENDFOR
\STATE /* Training */
\STATE Initialize policy model $\pi_\theta \leftarrow \pi_{\rm ref}$, mining coefficient $\beta$
\FOR{epoch in $1, 2, \cdots ,{\rm Epochs}$} 
    \STATE Update $\pi_\theta$ by Equation~(\ref{eqn:obj})
\ENDFOR
\STATE \textbf{Output:} $\pi_\theta$
\end{algorithmic}
\end{algorithm}

\section{Experiments}
\subsection{Experiment Setting}
\subsubsection{Training Dataset Construction} 
For the offline RL training dataset, we build the prompt seed set based on open-source data collected by the Open-Reasoner-Zero project \cite{hu2025open}, comprising AIME (up to 2023), OpenR1-Math-220k \cite{ben2025open}, and various other open-source datasets \cite{li2024numinamath, lambert2025tulu3pushingfrontiers, hendrycks2021measuring}. 
During data filtering, we exclude multiple-choice and true/false questions to prevent cases where the model might occasionally derive the correct answer through incorrect reasoning steps, which could result in false positives and introduce noise into the training process. 

Subsequently, we employ DeepSeek-R1-Distill-Qwen-14B (DS-R1-14B) to generate responses. For each question, we sample 32 responses to ensure sufficient coverage of the response space. We set the maximum response length as 22,000 tokens. The sampling temperature is set to 0.7 to maintain a balance between response diversity and quality. 

After generation, we use the open-source verifier tool \texttt{math\_verify} \cite{mathverify2025} to check whether the model's final answer is correct, and then remove questions for which all responses are either entirely correct or entirely incorrect, following \cite{light-r1}. After filtering, the statistics of our final offline RL dataset are summarized in Table \ref{data:stat}. 

\subsubsection{Evaluation Details}
For evaluation, we choose the highly challenging benchmarks AIME24 \cite{aime} and AIME25, along with MATH500 \cite{hendrycks2021measuring}, to demonstrate the model's mathematical reasoning performance. We also incorporate LiveCodeBench \cite{jain2024livecodebench}  (2024/8/1 - 2025/2/1) to show the generalization capabilities to coding tasks. Following DeepSeek-AI \cite{deepseek-r1}, long CoT models are commonly deployed with a sampling temperature. In our evaluation, we set the temperature to 0.7 (identical to the temperature used in rollout). We report the pass@1 averaged over 40 runs on AIME24 and AIME25. For MATH500 and LiveCodeBench, we average over 10 runs, as these benchmarks exhibit relatively small variance between test runs. This ensures statistical robustness and mitigates randomness in sampling, better reflecting the model's true capabilities.

\begin{table}[tb]
\centering
\begin{tabular}{@{}llll@{}}
\toprule
\begin{tabular}[c]{@{}l@{}}Question \\ Count\end{tabular} & \begin{tabular}[c]{@{}l@{}}Total \\ Samples\end{tabular} & \begin{tabular}[c]{@{}l@{}}Negative \\ Samples\end{tabular} & \begin{tabular}[c]{@{}l@{}}Total \\ Tokens\end{tabular} \\ \midrule
2069                                                      & 66208                                                    & 14896                                                       & 470M                                                    \\ \bottomrule
\end{tabular}
\caption{Statistics of the offline RL dataset.}
\vspace{-5pt}
\label{data:stat}
\end{table}

\subsubsection{Models}\label{sec:models}

\textbf{Base Model.} 
We initialize our training from the DS-R1-14B. The reasons for choosing it as the starting model are as follows: 1) DS-R1-14B has undergone large-scale SFT training and exhibits a stable long CoT pattern with frequent alternation between correct and incorrect steps, making it an ideal starting model to validate the effectiveness of NSA. 2) Based on R1 results \cite{deepseek-r1}, the 14B model achieves comparable performance to the 32B variant, offering more generalizable findings than the 7B model while being more computationally efficient than the 32B model.

\textbf{LLM Judger.} 
We employ Claude-3.7-Sonnet(thinking) \cite{claude-3.7}, one of the latest slow-thinking models, as the judger model. 

\textbf{PRM.} 
We utilize Qwen-Math-PRM-7B \cite{zhang2025lessons} as the PRM for step-wise annotation. This choice is motivated by its outstanding performance on the ProcessBench \cite{zheng2024processbench} Benchmark, demonstrating its robust capability in process evaluation tasks. We set the threshold $\lambda$ as 0.6.

\subsubsection{Baselines}

We choose the following methods as baselines:

\begin{itemize}
\item RFT \cite{rft}: It exclusively utilizes positive samples from the offline RL dataset and updates the model parameters through SFT loss.
\item DPO \cite{rafailov2023direct}: It is a prominent approach in offline RL settings and directly optimizes the preference objectives without explicit reward modeling. This method has demonstrated remarkable effectiveness in preference-based learning tasks. 
\item TOPR \cite{topr}: As an offline RL variant, it combines truncated importance sampling for negative samples with RFT-style optimization for positive samples, and removes KL regularization. 
\item GRPO-offline \cite{shao2024deepseekmath}: It applies the GRPO loss consistently throughout the offline RL training process, eliminating the need for periodic online data resampling. 
\end{itemize}

The training details of all algorithms can be found in Section \ref{sec:training_details}.

\subsection{Main Results}

\begin{table*}[ht]
\centering
\begin{tabular}{@{}cccccc|cc@{}}
\toprule
              & DS-R1-14B & RFT   & DPO   & TOPR  & GPRO-offline & BCPG           & BCPG-NSA       \\ \midrule
AIME24      & 70.58     & 62.92 & 69.83 & 69.75 & 70.50        & 70.50          & \textbf{72.17} \\
AIME25      & 49.58     & 50.75 & 49.00 & 52.67 & 50.90        & 52.00          & \textbf{54.42} \\
MATH500       & 91.80     & 92.20 & 92.88 & 93.12 & 92.08        & \textbf{93.98} & 93.36          \\
LiveCodeBench & 52.40     & 52.69 & 51.61 & 52.90 & 52.97        & 53.26          & \textbf{53.84} \\ \midrule
Average       & 66.09     & 64.64 & 65.83 & 67.11 & 66.61        & 67.44          & \textbf{68.45}          \\ \bottomrule
\end{tabular}
\caption{Evaluation results of BCPG-NSA and baselines on different benchmarks ($\beta$ is set to 0.5 for BCPG-NSA).}
\label{tab:main_results}
\end{table*}

\textbf{Negative samples play a crucial role in enhancing the performance of long CoT model.}
As shown in Table \ref{tab:main_results}, all offline RL methods consistently outperform RFT. Notably, RFT performs even worse than the original DS-R1-14B, exhibiting a significant 7.5\% drop on AIME24.
These results indicate that positive samples alone are insufficient to improve a distilled model that has already undergone SFT on a large-scale dataset. The negative gradient from negative samples can reduce the probability of wrong reasoning content, which is crucial for advancing model performance in long CoT reasoning.
We hypothesize that the reason for the decline in RFT performance is that a substantial proportion of questions in our training dataset have already been utilized in the distillation phase. For the same questions, the answers generated by DS-R1-14B are not as good as those generated by R1. Therefore, during RFT, the model memorizes the poorer responses and forgets the better ones from the distillation phase.

\textbf{Vanilla BCPG sets a competitive foundation.}
Despite its simplicity, BCPG achieves a 1.4\% average improvement across all benchmarks compared to DS-R1-14B, outperforming other offline RL methods. Given its superior performance, we select BCPG as the foundation to validate the effectiveness of NSA. 

\textbf{Negative sample augmentation leads to significant performance gain.}
Compared to BCPG, BCPG-NSA achieves a remarkable 2.5\% improvement on AIME25 and a 1\% improvement on average performance. Since DS-R1-14B already performs very well on the AIME24 benchmark, previous work \cite{light-r1} has shown that achieving further improvements on AIME24 is quite challenging. However, our BCPG-NSA algorithm successfully achieves an accuracy exceeding 72\%. 
These results validate that the correct steps within negative samples are of great value and can significantly enhance the model’s reasoning ability.

\section{Ablation Studies}
\subsection{Different Annotation Methods}
In this section, we investigate the impact of the annotation quality on the performance of BCPG-NSA. We conduct experiments where negative samples are annotated using PRM only and the LLM judger only, respectively, and compare with the performance obtained by consensus filtering that integrates both LLM and PRM annotations. Table \ref{tab:steps_cnt_statistic} shows the number of tokens in correct steps and incorrect steps among negative samples under different annotation methods. In all three experiments mentioned above, $\beta$ is set to 0.5. 

Results are demonstrated in Table \ref{tab:annotation_precision}.
Compared to the base model DS-R1-14B and the vanilla BCPG, negative sample augmentation consistently improves performance across all 3 annotation methods, demonstrating the robustness of NSA's benefits regardless of the choice of judgers. 
Moreover, the LLM-PRM approach, despite mining the smallest number of correct tokens (only 26M in total) from negative samples, achieves the best performance. We hypothesize that the consensus-based filtering implements more stringent selection criteria, ensuring a higher quality of retained correct steps and avoiding potential false positives from individual judgers.

\begin{table}[htbp]
\centering
\begin{tabular}{@{}ccc@{}}
\toprule
          & Correct tokens  & Incorrect tokens         \\ \midrule
PRM-only & 38M &  80M        \\
LLM-only    & 65M & 53M \\ 
LLM-PRM  & 26M &  92M     \\ 
\bottomrule
\end{tabular}
\caption{Number of tokens in correct/incorrect steps of negative samples under different annotation methods.}
\vspace{-5pt}
\label{tab:steps_cnt_statistic}
\end{table}

\begin{table}[htbp]
\centering
\begin{tabular}{@{}ccc|c@{}}
\toprule
          & AIME24 & AIME25 & Avg.            \\ \midrule
DS-R1-14B & 70.58  & 49.58  & 60.08          \\
BCPG      & 70.50  & 52.00  & 61.25          \\
PRM-only  & 70.58  & \textbf{55.58}  & 63.08          \\
LLM-only  & \textbf{72.17}  & 53.33  & 62.75          \\
LLM-PRM   & \textbf{72.17}  & 54.42  & \textbf{63.30} \\ \bottomrule
\end{tabular}
\caption{Performance under different annotation methods.}
\vspace{-5pt}
\label{tab:annotation_precision}
\end{table}

\subsection{Different Values of Mining Coefficient}
Mining coefficient $\beta$ is the key hyperparameter in NSA, used to control the strength of the augmentation. Therefore, we test the performance of the model under different $\beta$ values within the range $[-1, 1]$ to demonstrate the robustness of NSA. When $\beta=1$, BCPG-NSA reduces to the vanilla BCPG method.

\begin{figure}[tb]
    \centering
    \includegraphics[width=1\linewidth]{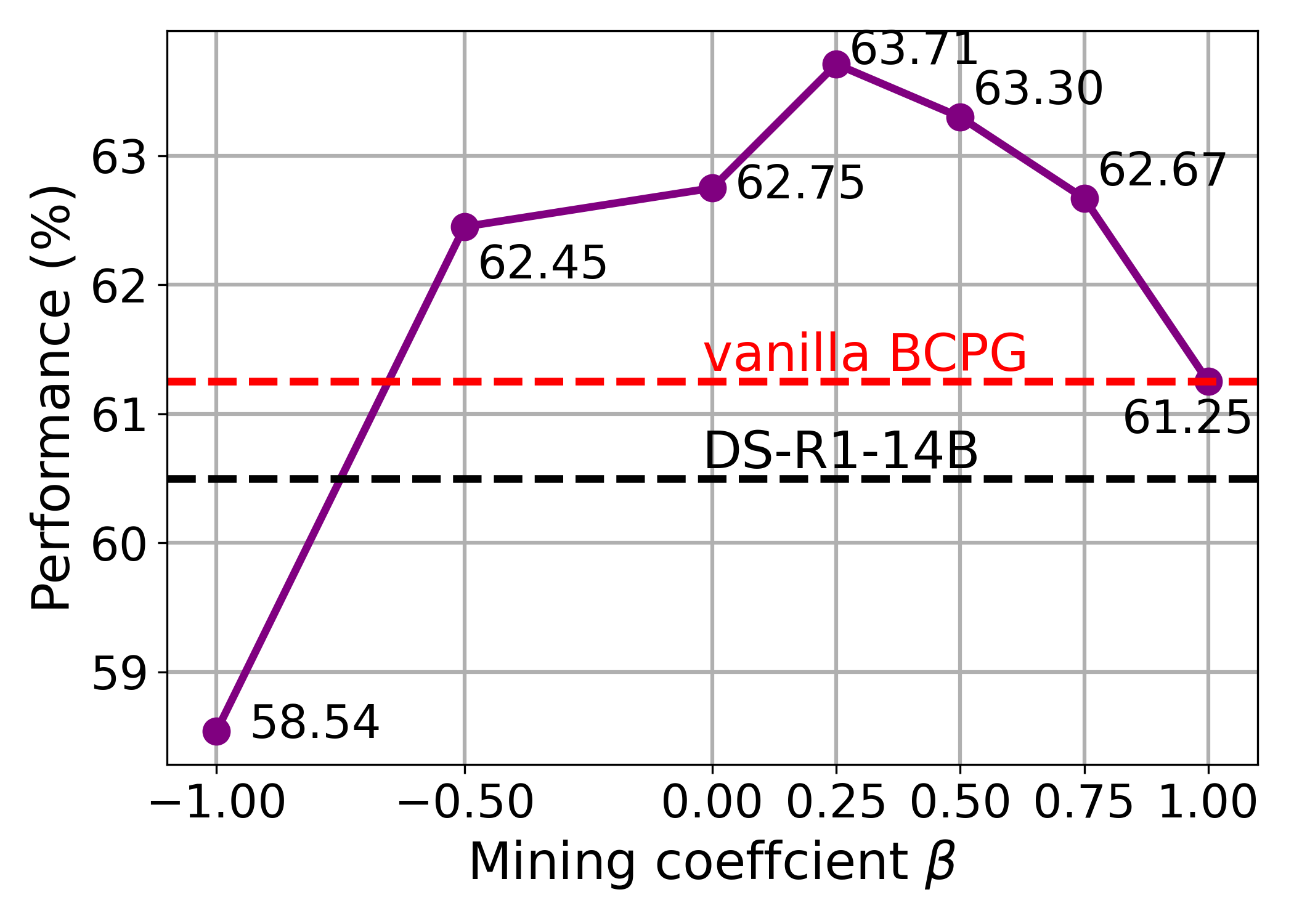}
    \vspace{-5pt}
    \caption{Average performance of AIME24 and AIME25 across different $\beta$ values.}
    \vspace{-5pt}
    \label{fig:beta_ablation}
\end{figure}

Results in Figure \ref{fig:beta_ablation} demonstrate that: 1) As $\beta$ decreases (indicating more aggressive negative sample augmentation), BCPG-NSA's performance exhibits an initial increase followed by a decline. 2) BCPG-NSA outperforms vanilla BCPG across a wide range of $\beta$ values, including at the relatively aggressive setting of $\beta=-0.5$. 
This robust performance across different $\beta$ values suggests that NSA is not overly sensitive to this hyperparameter and can consistently deliver performance improvements.

\section{Further Analysis}
\subsection{Training Dynamics}
\begin{figure}[tb]
    \centering
    \includegraphics[width=1\linewidth]{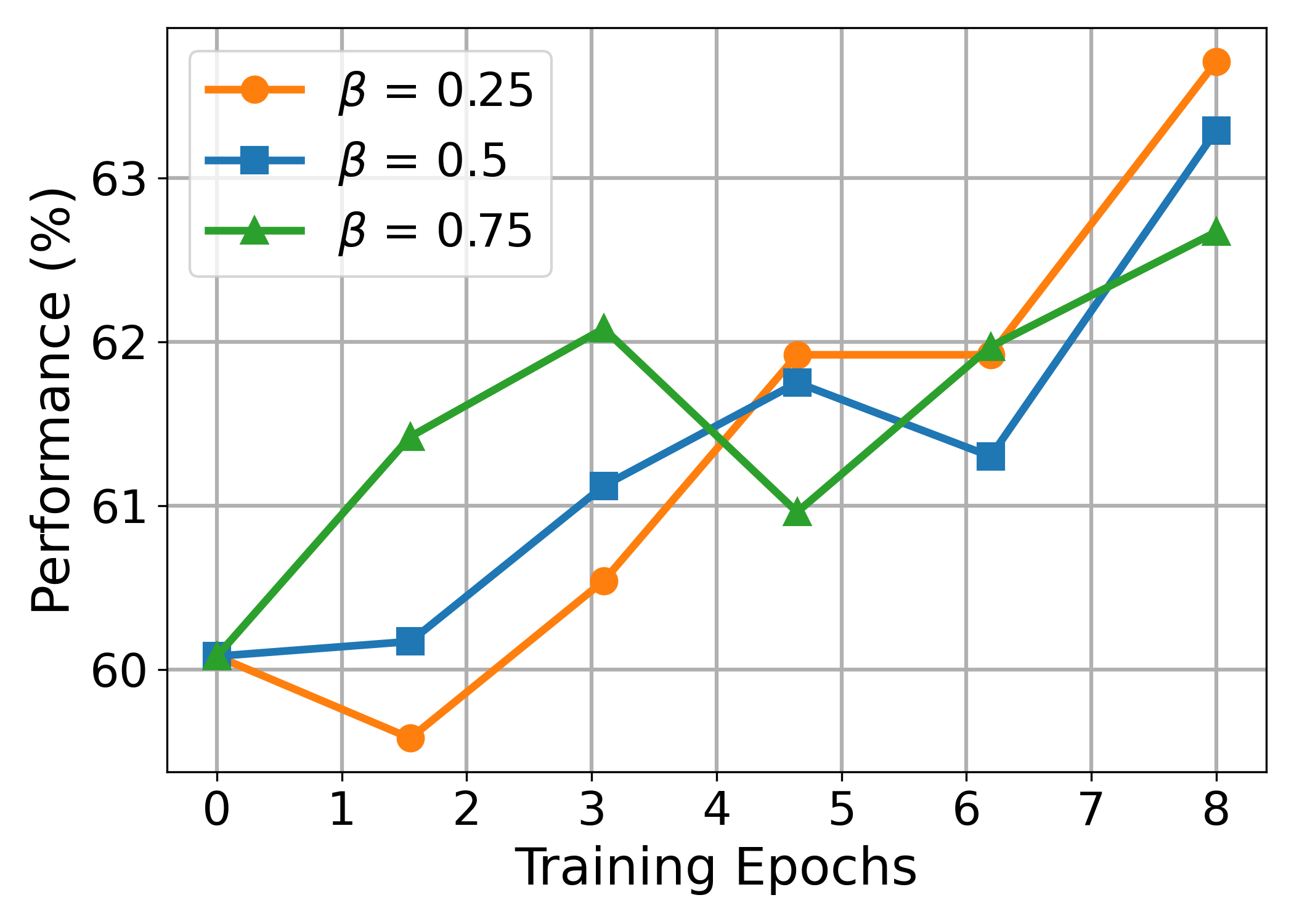}
    \vspace{-5pt}
    \caption{Average performance of AIME24 and AIME25 under different update steps (single iteration).}
    \label{fig:performance_with_steps}
\end{figure}

In offline RL, the data in the training set is often used multiple times for updates. Therefore, we track the average performance of AIME24 and AIME25 under different update steps during training. We conduct experiments with $\beta \in [0.25, 0.5, 0.75]$. The experimental results in Table \ref{fig:performance_with_steps} show that as the number of training epochs increases, the model’s performance improves steadily. Therefore, conducting multiple rounds of updates is essential to fully leverage the rollout data and achieve higher sample efficiency.
\subsection{The Performance of Multiple Iterations}
To investigate the scalability of our proposed method when extended to multiple training iterations, we further evaluate both BCPG-NSA and BCPG in a second iteration. Given the enhanced model capabilities after the first iteration, we sample 1,000 problems from the 13k hard dataset \cite{hu2025open} derived from the Open-Reasoner-Zero project for the second iteration. Using the model checkpoint from the end of iteration 1, we perform rollouts on these problems and generate 32 responses per problem.
As shown in Table \ref{tab:multi_iteration}, BCPG-NSA achieves an additional 1.3\% performance improvement on AIME25 in iteration 2, demonstrating its ability to continuously benefit from multiple iterations. Notably, BCPG's performance in iteration 2 remains below that of BCPG-NSA in iteration 1, further validating the effectiveness of NSA.

\begin{table}[htbp]
\centering
\resizebox{\columnwidth}{!}{
\begin{tabular}{@{}lcc|l@{}}
\toprule
                       & AIME24         & AIME25         & Avg.  \\ \midrule
DS-R1-14B              & 70.58          & 49.58          & 60.08 \\ \midrule
BCPG (iteration 1)     & 70.50          & 52.00          & 61.25 \\
BCPG-NSA (iteration 1) & \textbf{72.17} & 54.42          & 63.30 \\ \midrule
BCPG (iteration 2)     & 71.42          & 53.42          & 62.42 \\
BCPG-NSA (iteration 2) & 72.00          & \textbf{55.75} & \textbf{63.88} \\ \bottomrule
\end{tabular}
}
\caption{Performance under multiple iterations.}
\vspace{-5pt}
\label{tab:multi_iteration}
\end{table}

\section{Conclusion}

In this paper, we present BCPG-NSA, a novel offline RL framework designed to effectively leverage valuable components within negative samples for LLMs' long CoT reasoning. Unlike existing methods that either discard negative samples or apply equal penalization, our approach enables fine-grained treatment of negative samples through three key components: semantic step segmentation, consensus-based step correctness assessment, and policy optimization with negative sample augmentation.
Experimental results on challenging reasoning benchmarks demonstrate that BCPG-NSA achieves superior performance and exhibits great robustness and scalability when extended to multiple iterations.

\section{Limitations}
Although our proposed negative sample augmentation framework has demonstrated its effectiveness on multiple challenging mathematical and code reasoning benchmarks, there are still several limitations to be addressed. 
1) More fine-grained credit assignment. Currently, our method assigns equal credit to all correct steps in negative samples. A potential improvement would be to enable LLM or PRM judger to output continuous values instead of binary decisions, which could then be used to compute a weighted final score, allowing for more precise credit assignment for each step. 
2) While the consensus-based LLM-PRM annotation method achieves strong performance, the annotation process could be optimized by training a dedicated model on the labeled data, potentially improving both efficiency and speed. These improvements will be explored in future work.
\section{Acknowledgments}
This work was supported in part by the Strategic Priority Research Program of Chinese Academy of Sciences under Grant XDA0480301.
\bibliography{main}

\newpage

\appendix

\section{Case Study: A negative response that incorporates positive steps}
\label{sec:case}
 Figure \ref{fig:case-part1} and Figure \ref{fig:case-part2} present a detailed case to more specifically demonstrate the results of segmentation and annotation of the negative samples. Model's response is segmented into 15 steps. The \textit{content} field contains the model's reasoning process in each step. A \textit{score} of 1 and 0 indicate the step is correct and incorrect, respectively. The \textit{reason} field provides the rationale for the scoring.

\begin{figure*}[ht]
\begin{tcolorbox}[colframe=black, colback=lightblue, coltitle=black, rounded corners, boxrule=0.3mm, width=\textwidth, halign=left]

\textbf{Problem:} What is the sum of all of the multiples of 7 between 100 and 200?

\textbf{Solution:}
\begin{itemize}
    \item Step 1
    \begin{itemize}
        \item \textbf{Content:} Hmm, okay. Let me think about this step by step. Maybe I can start by recalling how to find the sum of an arithmetic series because multiples of 7 should form an arithmetic sequence. Wait, arithmetic series? Yes, because each term increases by a constant difference, which in this case is 7. So, I can use the formula for the sum of an arithmetic series. ... Then, dividing both sides by 7, I get 13 = n - 1. Adding 1 to both sides, n = 14. So, there are 14 terms in this sequence.
        \item \textbf{Score:} 1.
        \item \textbf{Reason:} Correctly identifies the first term (105), last term (196), and number of terms (14) in the arithmetic sequence.
    \end{itemize}
    
    \item Step 2
    \begin{itemize}
        \item \textbf{Content:} Wait, let me verify that. Let's list out some terms to be sure. Starting at 105: 105, 112, 119, 126, 133, 140, 147, 154, 161, 168, 175, 182, 189, 196. Yeah, that's 14 terms. Okay, so n = 14. Now, plugging these values into the sum formula: $S = n/2*(\mathrm{first\ term} + \mathrm{last\ term})$. So, that would be $S = 14/2*(105+196)$. Calculating that, 14 divided by 2 is 7, and 105 plus 196 is 299. So, 7 multiplied by 299.Hmm, 7*300 is 2100, so 7*299 is 2100 minus 7, which is 2093.Wait, really? ...
        \item \textbf{Score:} 0.
        \item \textbf{Reason:} Miscalculates the sum by incorrectly adding the first and last terms (105 + 196 = 299 instead of 301) and incorrectly computes the sum up to 200 using the wrong first term (105 instead of 7).
    \end{itemize}
    \item Step 3
    \begin{itemize}
        \item \textbf{Content:} Wait, that's a different number. So, which one is correct? ... Alternatively, since all these numbers form an arithmetic progression, I can use the formula again. So, $S = n/2*(\mathrm{first\ term} + \mathrm{last\ term})$. So, n is 14, first is 105, last is 196. So, 14/2 is 7, times (105 + 196) is 7*299=2093. ...
        \item \textbf{Score:} 0.
        \item \textbf{Reason:} Continues based on STEP 2’s errors (299 instead of 301 and incorrect sum up to 200) without correction or a new correct approach.
    \end{itemize}
    \item Step 4
    \begin{itemize}
        \item \textbf{Content:} ... Wait, maybe I miscalculated one of the sums. Let me compute the sum from 105 to 196 manually, but that would be time-consuming, but let me try a few terms. 105 + 112 + 119 + 126 + 133 + 140 + 147 + 154 + 161 + 168 + 175 + 182 + 189 + 196. Let me add them step by step. Start with 105. 105 + 112 = 217. 217 + 119 = 336. 336 + 126 = 462. 462 + 133 = 595. 595 + 140 = 735. 735 + 147 = 882. 882 + 154 = 1036. 1036 + 161 = 1197. 1197 + 168 = 1365. 1365 + 175 = 1540. 1540 + 182 = 1722. 1722 + 189 = 1911. 1911 + 196 = 2107.
        \item \textbf{Score:} 1.
        \item \textbf{Reason:} Correctly verifies the sum as 2107 through manual addition and the difference method, despite repeating an earlier incorrect calculation.
    \end{itemize}
\end{itemize}

\end{tcolorbox}
\caption{An example of negative response that incorporates positive steps (part 1).}
\label{fig:case-part1}
\end{figure*}

\begin{figure*}[ht]
\begin{tcolorbox}[colframe=black, colback=lightblue, coltitle=black, rounded corners, boxrule=0.3mm, width=\textwidth, halign=left]

\begin{itemize}
    \item Step 5
    \begin{itemize}
        \item \textbf{Content:} ...... Wait, so the manual addition gives me 2107, same as 2842 - 735.But when I used the arithmetic series formula, I got 2093. So, that's the problem. Why is there a discrepancy? Wait, let me check the arithmetic series formula again. I think the issue is with the number of terms.Wait, the first term is 105, which is 15*7, and the last term is 196, which is 28*7.
        \item \textbf{Score:} 1.
        \item \textbf{Reason:} Correctly applies the alternative arithmetic series formula to get 2107, confirming the correct sum.
    \end{itemize}
    \item Step 6
    \begin{itemize}
        \item \textbf{Content:} Wait, earlier when I did the arithmetic series formula correctly, I think I messed up. Wait, no. If I use $S = n/2 * (\mathrm{first} + \mathrm{last})$, which is $(14/2 \times (105+196)=7\times 299=2093$. But when we I used the correct formula $S = n/2 * (2a+(n-1)d)$, it gave me 2107. Wait, that's a contradiction. So, what's happening here. Wait, perhaps I made an error in the number of terms. ... So, n=14, first term 105, last term 196. So, the sum is $14/2*(105+196)=7*299=2093$ ...
        \item \textbf{Score:} 0.
        \item \textbf{Reason:} Reintroduces the incorrect sum (2093) alongside the correct 2107 without resolving the error.
    \end{itemize}
    \item Step 7-14. A series of deductions based on false premises.
    \item Step 15
        \begin{itemize}
        \item \textbf{Content:} So, there are 14 terms in this sequence. We then use the arithmetic series sum formula: $S = \frac{n}{2} \times (\mathrm{first \ term} + \mathrm{last \ term})$. Substituting the values, we get: $S=\frac{14}{2}\times(105+196)=7\times299=2093$. Thus, the sum of all multiples of 7 between 100 and 200 is $\boxed{2093}$.
        \item \textbf{Score:} 0.
        \item \textbf{Reason:} Concludes with the incorrect sum of 2093 instead of the correct 2107.
    \end{itemize}
\end{itemize}
\end{tcolorbox}
\caption{An example of negative response that incorporates positive steps (part 2).}
\label{fig:case-part2}
\end{figure*}

\section{Prompt Template for LLM Judger}
\label{sec:Prompt Template for LLM Judger}
The prompt template for LLM judger in Section \ref{sec:models} is shown in Figure \ref{fig:prompt_template_judger}.
\begin{figure*}[ht]
\begin{tcolorbox}[colframe=black, colback=lightblue, coltitle=black, rounded corners, boxrule=0.3mm, width=\textwidth, halign=left]
You are an expert in evaluating mathematical problem-solving processes. The user will provide a math problem, a step-by-step solution, and ground truth. Your tasks are as follows:
\begin{enumerate}
    \item First, extract a concise final answer (Short Answer) from the ground truth.
    \item Then, carefully review the user's step-by-step solution, assigning a score to each step (either 0 or 1). For each step, provide a brief explanation of your judgment result.
\end{enumerate}
Scoring rules:
\begin{itemize}
    \item If a step contains an explicit error, such as a reasoning error, calculation mistake, logical flaw, or misunderstanding of the problem, it should be scored 0.
    \item If a step does not contain any errors, score it according to the following rules:
    \begin{enumerate}  
    \item \textbf{Error Propagation:} If a previous step contains an error and the current step continues the analysis based on that error without introducing a new, correct approach or making a proper correction, the current step should also be scored 0.
    \item \textbf{Error Termination:} If a previous step contains an error, but the current step either corrects the previous mistake or introduces a new and correct approach, the current step should be scored 1. For example:
    \begin{itemize}
        \item STEP K contains an error.
        \item STEP K+1 continues the analysis based on the error.
        \item STEP K+2 corrects the previous error or introduces a new and correct approach. 
    \end{itemize}
    In this case, STEP K and STEP K+1 should be scored 0, and STEP K+2 should be scored 1.
    \end{enumerate}
\end{itemize}

Your response format should be in json format: \texttt{\newline 
\phantom{AAAA}[\newline
\phantom{AAAAAAAA}\{ \newline
\phantom{AAAAAAAAAAAA}"STEP 0": 1(int), \newline
\phantom{AAAAAAAAAAAA}"Reason": xxxx(str) \newline
\phantom{AAAAAAAA}\}, \newline
\phantom{AAAAAAAA}\{ \newline
\phantom{AAAAAAAAAAAA} "STEP 1": 1(int), \newline
\phantom{AAAAAAAAAAAA} "Reason": xxxx(str)  \newline
\phantom{AAAAAAAA}\} \newline
\phantom{AAAAAAAA}... \newline
\phantom{AAAA}] \newline}

\textbf{Note:} When analyzing the solution, remain objective and rational. Do not be misled by the way the user's solution is described.

\end{tcolorbox}
\caption{Prompt template for the LLM judger.}
\label{fig:prompt_template_judger}
\end{figure*}

\section{Training Details}
\label{sec:training_details}
The detailed hyperparameters of the RFT method are shown in Table \ref{tab:parameters_of_rft}.

\begin{table}[h]
\centering
\begin{tabular}{@{}l|c@{}}
\toprule
Parameter         & Value                \\ \midrule
Max Learning Rate & $2.5 \times 10^{-6}$ \\
Min Learning Rate & $1 \times 10^{-7}$   \\
Warmup Fraction   & 0.01                 \\
Epochs            & 6                    \\
Batch Size        & 64                   \\
Seq Length        & 32k                  \\ \bottomrule
\end{tabular}
\caption{Training hyperparameters of RFT.}
\label{tab:parameters_of_rft}
\end{table}

For other RL methods, the shared hyperparameter settings are shown in Table \ref{tab:parameters_of_other_methods}. All experiments are conducted using 64 NVIDIA H800/H100 GPUs. The training of BCPG-NSA takes approximately 14 hours.
\begin{table}[h]
\centering
\begin{tabular}{@{}l|c@{}}
\toprule
Parameter         & Value                \\ \midrule
Max Learning Rate & $5 \times 10^{-7}$   \\
Min Learning Rate & $2.5 \times 10^{-7}$ \\
Epochs            & 8                    \\
Batch Size        & 512                  \\
Seq Length        & 32k                  \\ \bottomrule
\end{tabular}
\caption{Training hyperparameters of offline RL methods.}
\label{tab:parameters_of_other_methods}
\end{table}

For both the BCPG and BCPG-NSA, $\tau$ is set to $1\times 10^{-3}$. For the DPO algorithm (Equation \ref{equation:dpo}), we set $\beta_{\rm DPO}$ to 0.5.
\begin{align}
    \label{equation:dpo}
    J&_{\mathrm{DPO}}  (\pi_{\theta} ;\pi_{\mathrm{ref}}) = - \mathbb{E}_{(x,y_w,y_l)\sim D}\bigg[ \notag \\
    & \log \sigma \left( \beta_{\rm DPO} \left( \log\frac{\pi_{\theta}(y_w\mid x)}{\pi_\mathrm{ref}(y_w\mid x)} \right. \right. \\
    & \phantom{\text{xxxxxxxxxxxx}} \left. \left. - \log\frac{\pi_{\theta}(y_l\mid x)}{\pi_\mathrm{ref}(y_l\mid x)} \right) \right) \bigg] \notag
\end{align}

For the objective of TOPR algorithm (Equation \ref{equation:topr}), the upper bound of the clip $a$ is set to 1, and the lower bound $b$ is set to 0.

\begin{equation}
    \label{equation:topr}
    J_{\mathrm{TOPR}}(\pi_{\theta}) = - \mathbb{E}_{y\sim \pi_{\mathrm{ref}}} \bigg[ \rho(x,y) \ \mathrm{log}\pi_{\theta}(y|x)\bigg].
\end{equation}

In the equation above, $\rho(x,y)$ is defined as:
\begin{equation}
\rho(x,y) = \begin{cases}
\left[\frac{\pi_{\theta}(y|x)}{\pi_{\mathrm{ref}}(y|x)}\right]^a_b, & \text{if } r(x,y,y^*) < 0 \\
1, & \text{otherwise}.
\end{cases}
\end{equation}

For the GRPO algorithm (Equation \ref{equation:grpo}), we set $\epsilon = 0.2$ and $\beta_{\rm GRPO} = 1 \times 10^{-3}$.

\begin{align}
    \label{equation:grpo}
    J_{\mathrm{GRPO}}(\pi_{\theta}) & = - \mathbb{E}_{x\sim D, \{y_i\}_{i=1}^G\sim \pi_{\mathrm{ref}}} \bigg[ \notag \\ & \frac{1}{G}\sum_{i=1}^G\bigg( \mathrm{min} \big(\frac{\pi_{\theta}(y_i|x)}{\pi_{\mathrm{ref}}(y_i|x)}A_i, \notag \\ & \mathrm{clip}(\frac{\pi_{\theta}(y_i|x)}{\pi_{\mathrm{ref}}(y_i|x)}), 1-\epsilon, 1+\epsilon \big)A_i\bigg) \notag \\ & -\beta_{\rm GRPO} D_{KL}(\pi_{\theta}||\pi_{\mathrm{ref}})\bigg]
\end{align}

\end{document}